\documentclass[10pt,twocolumn,letterpaper]{article}

\usepackage{iccv}
\usepackage{epsfig}
\usepackage{amsmath}
\usepackage{amssymb}
\usepackage{times,graphicx,tabulary,multirow,xspace,xcolor}
\usepackage[inline, shortlabels]{enumitem}
\usepackage{booktabs}
\usepackage{color, colortbl}

\usepackage{pifont}
\newcommand{\cmark}{\color{gray}\ding{51}}%
\newcommand{\xmark}{\color{gray}\ding{55}}%
\definecolor{darkgreen}{rgb}{0.29, 0.33, 0.13}
\definecolor{darkred}{rgb}{0.75, 0.0, 0.0}
\definecolor{darkblue}{rgb}{0.0, 0.0, 0.75}
\definecolor{LightGray}{rgb}{0.92,0.92,0.92}

\newcommand{\eat}[1]{}

\usepackage[pagebackref=true,breaklinks=true,letterpaper=true,colorlinks,bookmarks=false]{hyperref}

\iccvfinalcopy 

\def\iccvPaperID{7105} 

\ificcvfinal\fi
\begin{document}

\title{SAT: 2D Semantics Assisted Training for 3D Visual Grounding}

\author{Zhengyuan Yang$^1$ \quad Songyang Zhang$^1$ \quad Liwei Wang$^2$ \quad Jiebo Luo$^1$\\ $^1$University of Rochester \qquad\qquad $^2$The Chinese University of Hong Kong \\ 
{\tt\small \{zyang39, szhang83, jluo\}@cs.rochester.edu \quad lwwang@cse.cuhk.edu.hk}
}

\maketitle
\ificcvfinal\thispagestyle{empty}\fi

\begin{abstract}
  3D visual grounding aims at grounding a natural language description about a 3D scene, usually represented in the form of 3D point clouds, to the targeted object region. Point clouds are sparse, noisy, and contain limited semantic information compared with 2D images. These inherent limitations make the 3D visual grounding problem more challenging. In this study, we propose 2D Semantics Assisted Training (SAT) that utilizes 2D image semantics in the training stage to ease point-cloud-language joint representation learning and assist 3D visual grounding. The main idea is to learn auxiliary alignments between rich, clean 2D object representations and the corresponding objects or mentioned entities in 3D scenes. SAT takes 2D object semantics, \ie, object label, image feature, and 2D geometric feature, as the extra input in training but does not require such inputs during inference.
By effectively utilizing 2D semantics in training, our approach boosts the accuracy on the Nr3D dataset from $37.7\%$ to $49.2\%$, which significantly surpasses the non-SAT baseline with the identical network architecture and inference input. 
Our approach outperforms the state of the art by large margins on multiple 3D visual grounding datasets, \ie, $+10.4\%$ absolute accuracy on Nr3D, $+9.9\%$ on Sr3D, and $+5.6\%$ on ScanRef.
Code is available at \href{https://github.com/zyang-ur/SAT}{https://github.com/zyang-ur/SAT}.
\end{abstract}

\vspace{-8pt}
\section{Introduction}
\begin{figure}[t]
\begin{center}
  \centerline{\includegraphics[width=8.5cm]{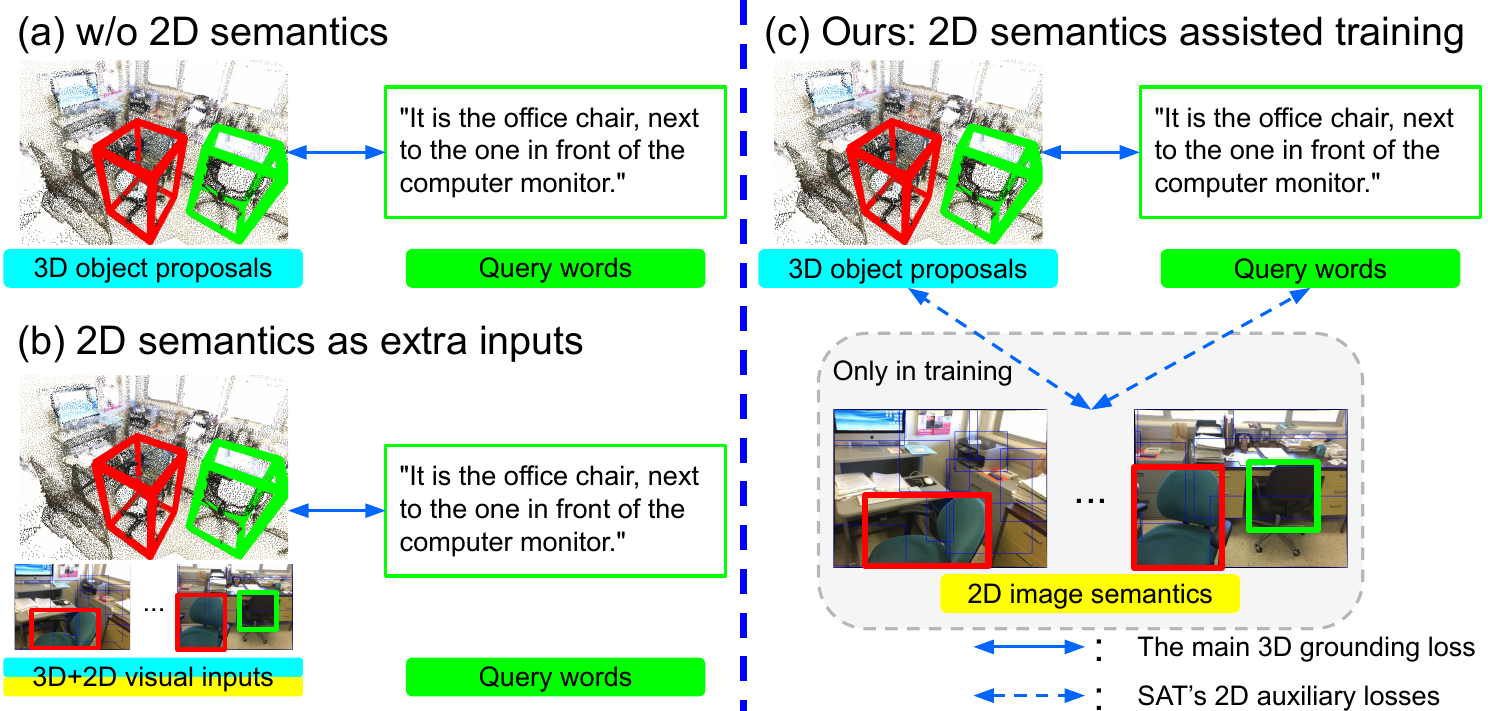}}
\end{center}
\vspace{-0.2in}
    \caption{
    3D visual grounding aims to ground a language query to a targeted 3D object region, as shown by the \textcolor{darkgreen}{green} 3D bounding box.
    \textbf{(a)} 
    Previous 3D visual grounding studies are trained with a sole 3D grounding loss that maximizes the similarity between positive object-query pairs. However, the sole objective is less effective as point clouds are sparse and noisy. \textbf{(b)} 2D semantics contain rich and clean object representations and can be used as extra visual inputs to assist 3D grounding. However, requiring extra 2D inputs in inference limits potential application scenarios. \textbf{(c)} Our proposed \textit{2D Semantics Assisted Training (SAT)} uses 2D semantics only in training and does not require extra inputs in inference.
    The \textcolor{darkgreen}{green} and \textcolor{darkred}{red} boxes are the targeted and distracting objects.
	}
\vspace{-0.15in}
\label{fig:intro}
\end{figure}

Visual grounding provides machines the ability to ground a language description to the targeted visual region. The task has received wide attention in both datasets~\cite{yu2016modeling,plummer2017flickr30k,kazemzadeh2014referitgame} and methods~\cite{hu2016natural,wang2016learning,yu2018mattnet,yang2019fast}. However, most previous visual grounding studies remain on images~\cite{yu2016modeling,plummer2017flickr30k,kazemzadeh2014referitgame} and videos~\cite{zhang2020does,sadhu2020video,yang2020grounding}, which contain 2D projections of inherently 3D visual scenes.
The recently proposed 3D visual grounding task~\cite{achlioptas2020referit3d,chen2019scanrefer} aims to ground a natural language description about a 3D scene to the region referred to by a language query (in the form of a 3D bounding box). The 3D visual grounding task has various applications, including autonomous agents~\cite{savva2019habitat,xia2018gibson}, human-machine interaction in augmented/mixed reality~\cite{kim2018revisiting,kress201711}, intelligent vehicles~\cite{mittal2020attngrounder,feng2021cityflow}, and so on.

Visual grounding tries to learn a good joint representation between visual and text modalities, \ie, the 3D point cloud and language query in 3D visual grounding. As shown in Figure~\ref{fig:intro}~(a), previous 3D grounding studies~\cite{achlioptas2020referit3d,chen2019scanrefer,huang2021text,yuan2021instancerefer} directly learn the joint representation with a sole 3D grounding objective of maximizing the positive object-query pairs' similarity scores. Specifically, the model first generates a set of 3D object proposals and then fuses each proposal with the language query to predict a similarity score. The framework is trained with a sole objective that maximizes the paired object-query scores and minimizes the unpaired ones' scores.
However, direct joint representation learning is challenging and less effective since 3D point clouds are inherently sparse, noisy, and contain limited semantic information. Given that the 2D object representation provides rich and clean semantics, we explore using 2D image semantics to help 3D visual grounding.

How to assist 3D tasks with 2D image semantics remains an open problem. 
Previous studies on 3D object detection and segmentation have proposed a series of methods that take 2D semantics as extra visual inputs to assist 3D tasks. Representative approaches include aligning the 2D object detection results with 3D bounding boxes~\cite{qi2018frustum,xu2018pointfusion,lahoud20172d} and concatenating the image visual feature with 3D points~\cite{ku2018joint,hou20193d,wang2019densefusion,song2016deep,qi2020imvotenet}.
However, these methods require extra 2D inputs in both training and inference. The necessity of extra input 2D data during inference limits potential application scenarios since 2D inputs might not exist in inference or require extra pre-processing, such as 2D-3D matching and 2D detection. Instead of as extra visual inputs in both training and inference (as shown in Figure~\ref{fig:intro}~(b)), we explore using 2D semantics only in training to assist 3D visual grounding.

In this study, we propose \textit{2D Semantics Assisted Training} (SAT), which utilizes 2D image semantics (in the form of object label, image feature, and 2D geometric feature) to ease joint representation learning between the 3D scene and language query. 
As shown in Figure~\ref{fig:intro}~(c), in addition to the main 3D visual grounding loss~\cite{achlioptas2020referit3d,chen2019scanrefer} that maximizes the score between the paired 3D object and language query, SAT introduces auxiliary loss functions that align objects in 2D images with the corresponding ones in 3D point clouds or language queries\eat{respectively}. The learned auxiliary alignments effectively distill the rich and clean 2D object representation to assist 3D visual grounding.
Specifically, in SAT, we study the training loss design for auxiliary alignments and the encoding method for 2D semantics features. For the former, we propose an object correspondence loss based on the triplet loss~\cite{karpathy2015deep,faghri2017vse++,wang2019learning,li2019visual} for 3D and 2D object alignment. For the latter, we propose a transformer attention mask that generates good 2D semantics features and prevents leaking 2D inputs to the output module.

We experiment with the SAT approach on a transformer-based model~\cite{vaswani2017attention} we propose and name as 3D grounding transformer. We benchmark SAT on the Nr3D~\cite{achlioptas2020referit3d}, Sr3D~\cite{achlioptas2020referit3d}, and ScanRef~\cite{chen2019scanrefer} datasets.
The extra 2D semantics, together with SAT's specially designed way of using them, effectively help the model learn a better 3D object point cloud representation and ease joint representation learning.
With the same network architecture and inference inputs, SAT improves the grounding accuracy on Nr3D from the non-SAT baseline's $37.7\%$ to $49.2\%$.

In summary, our main contributions are:
\vspace{-6pt}
\begin{itemize} 
\setlength\itemsep{-3pt}
\item We propose \textit{2D Semantics Assisted Training} (SAT) that assists 3D visual grounding with 2D semantics. 
To the best of our knowledge, SAT {is the first method that }helps 3D tasks with 2D semantics in training but does not require 2D inputs during inference.
\item With the proposed object correspondence loss and the 2D semantics encoding method, SAT effectively utilizes 2D semantics to learn a better 3D object representation, which leads to significant \eat{grounding }accuracy improvements on the Nr3D~\cite{achlioptas2020referit3d}~($+10.4\%$ in absolute accuracy), Sr3D~\cite{achlioptas2020referit3d}~($+9.9\%$), and ScanRef~\cite{chen2019scanrefer}~($+5.6\%$) datasets.

\vspace{-3pt}
\end{itemize}
\section{Related Work}
\noindent\textbf{3D visual grounding.}
3D visual grounding aims to ground the language referred object in a 3D scene (in the form of RGB-XYZ point clouds) to a 3D bounding box. Two recent works Referit3D~\cite{achlioptas2020referit3d} and ScanRef~\cite{chen2019scanrefer}, independently proposed datasets and baseline methods for the 3D visual grounding task. Both works~\cite{achlioptas2020referit3d,chen2019scanrefer} augment the 3D scans in the ScanNet~\cite{dai2017scannet} dataset with the manually annotated language queries to construct the 3D visual grounding datasets. Previous 3D grounding studies~\cite{achlioptas2020referit3d,chen2019scanrefer,huang2021text,yuan2021instancerefer,feng2021free,roh2021languagerefer,zhao2021transrefer3d} follow a two-stage framework. In the first stage, multiple 3D object proposals are generated either with ground truth objects~\cite{achlioptas2020referit3d} or a 3D object detector~\cite{chen2019scanrefer,qi2019deep}. In the second stage, 3D object proposal features are fused with the language query to predict each proposal's matching scores. A softmax grounding loss is applied to maximize the score between the paired object proposal and language query.

We find that the sole objective of similarity score maximization is less effective because the point clouds for object proposals are sparse and noisy. In this study, we explore using 2D image semantics to assist 3D visual grounding.

\noindent\textbf{2D semantics in 3D tasks.}
Studies on 3D object detection and segmentation have explored using 2D image semantics to assist 3D tasks. There exist two representative approaches, \ie, 1) projecting image object detection results into 3D space to assist 3D box prediction~\cite{qi2018frustum,xu2018pointfusion,lahoud20172d} and 2) concatenating the image feature with each point in the 3D scene as the extra information for the 3D tasks~\cite{ku2018joint,hou20193d,wang2019densefusion,song2016deep,qi2020imvotenet}. ImVoteNet~\cite{qi2020imvotenet} fuses the image object detection results with 3D points.

Previous studies use 2D image semantics as the extra inputs to 3D tasks and thus require the extra 2D information in both training and inference. Despite the performance improvement, the extra 2D inputs potentially limit the application scenarios since the 2D information either does not exist in inference or requires tedious pre-processing, such as 2D-3D matching and 2D object detection. In this study, we explore using 2D semantics only in training to assist 3D visual grounding.

\noindent\textbf{Image visual grounding.}
3D visual grounding is related to the image visual grounding task~\cite{kazemzadeh2014referitgame,plummer2015flickr30k,yu2016modeling,mao2016generation}. There are mainly two approaches in image visual grounding, namely the one- and two-stage frameworks. The one-stage methods~\cite{yang2019fast,sadhu2019zero,yang2020improving,deng2021transvg} fuse the language query with each pixel/patch in image and predict grounding boxes densely at all spatial locations. The two-stage methods~\cite{yu2016modeling,wang2019learning,yu2018mattnet} first generate object proposals based on the visual objectiveness. The methods then compare each proposal with the language query to select the grounding prediction.

We follow previous 3D grounding studies~\cite{achlioptas2020referit3d,chen2019scanrefer,huang2021text} and experiment with our proposed SAT on a two-stage framework introduced in Section~\ref{sec:trans}. We focus on using 2D semantics to assist 3D grounding in this study and leave the exploration of alternative frameworks to future studies.
\section{3D Grounding Transformer}
\label{sec:trans}
\begin{figure*}[t]
\begin{center}
  \centerline{\includegraphics[width=16cm]{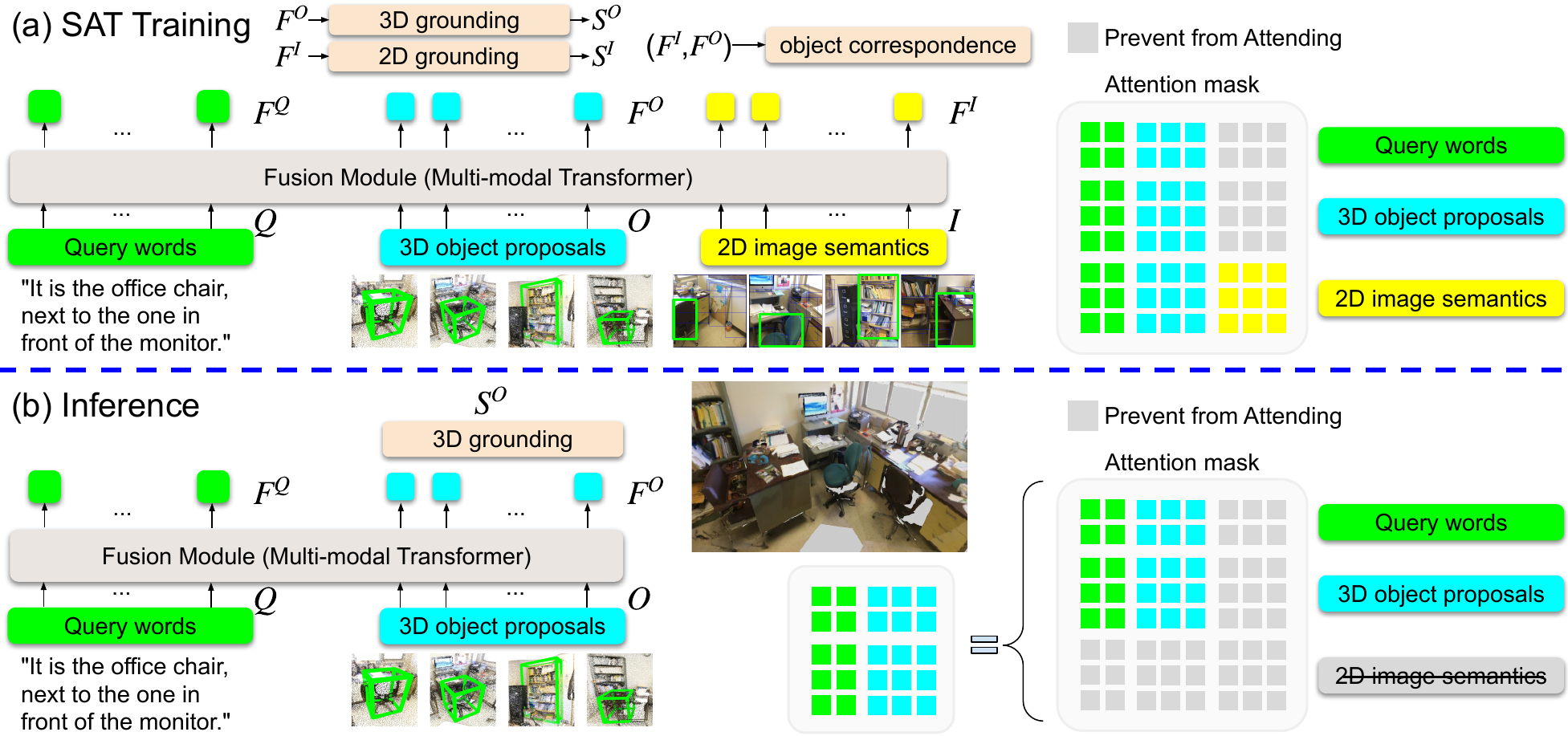}}
\end{center}
\vspace{-0.2in}
    \caption{The proposed 2D semantics assisted training (SAT) for 3D visual grounding. \textbf{(a)} In training, SAT takes 2D semantics as extra input and helps 3D visual grounding with the auxiliary objectives of 2D visual grounding $\mathcal{L}_{VG}^{I}$ and object correspondence prediction $\mathcal{L}_{cor}$. 
    \textbf{(b)} In inference, SAT does not require 2D inputs and is easy to use.
    SAT's attention mask prevents query words and 3D proposals from attending on 2D image semantics $I$ in training (top five rows of the mask), avoiding performance drop in inference when $I$ is not available.
	}
 \vspace{-0.1in}
\label{fig:arch}
\end{figure*}

Before introducing our proposed 2D semantics assisted training (SAT), we first overview the problem modeling of the 3D visual grounding task, and a transformer-based network architecture that we experiment on, named the 3D grounding transformer.

\subsection{3D visual grounding inputs}
The input to the 3D visual grounding task is a 3D scene $S\in \mathbb{R}^{N\times6}$ in the form of RGB-XYZ point clouds with $N$ points and a natural language query with $K$ words.
In the training stage, SAT takes the extra input of 2D semantics extracted from the original ScanNet videos~\cite{dai2017scannet} to ease joint representation learning. We detail how SAT represents and utilizes 2D semantics in following sections.

\subsection{Embedding for all modalities}
\label{sec:emb}
\noindent\textbf{3D scene embedding.}
Following previous studies~\cite{achlioptas2020referit3d,chen2019scanrefer,huang2021text}, we assume the access to $M$ 3D object proposals (in the form of point cloud object segments) in scene $S$. The proposals are either generated with ground truth objects as in Referit3D~\cite{achlioptas2020referit3d} or by a detection network~\cite{qi2019deep} as in ScanRef~\cite{chen2019scanrefer}. After getting the proposals, we normalize each object's center and size~\cite{wald2020learning}, and encode the point cloud segment of each proposal into a feature vector $x_m^{pc}$ with PointNet++~\cite{qi2017pointnet++,achlioptas2020referit3d,chen2019scanrefer,wald2020learning}. We obtain the $d$-dimensional 3D proposal embedding $\{O_1,\cdots,O_M\}$ with two learned linear transforms, where
\begin{equation*}
    O_m = \text{LN}(W_1 x_m^{pc})+\text{LN}(W_2 x_m^{\text{offset}}).
\end{equation*}
$x_m^{pc}$ is the PointNet++'s output feature. $x_m^{\text{offset}}$ is a 4D vector with the normalization offset, \ie, the center offsets $(x,y,z)$ and the original size $r$ for proposal $m$. $W_1$, $W_2$ are learned projection matrices. $\text{LN}(\cdot)$ is layer normalization~\cite{ba2016layer}.

\noindent\textbf{2D semantics embedding.}
For each 3D proposal $m$, we project its point clouds onto $L$ sampled frames in the original ScanNet videos~\cite{dai2017scannet} and get the corresponded 2D image semantics (image region, 2D bounding box, object class). In sampled frame $l\in \{1,\cdots,L \}$, we represent the 2D semantics for proposal $m$ by its visual feature $x_{m,l}^{ROI}$ (Region of interest feature from a visual genome~\cite{krishna2017visual} pre-trained Faster-RCNN detector~\cite{salvador2016faster}), semantic feature $x_{m,l}^{cls}$ (one-hot class vector), and geometric feature $x_{m,l}^{geo}$ (2D bounding box coordinates and frame's camera pose). We obtain the $d$-dimensional 2D semantics $I_{m,l}$ with linear transforms:
\begin{equation}
\label{eq:i}
    I_{m,l} = \text{LN}(W_3 x_{m,l}^{ROI}+W_4 x_{m,l}^{cls})+\text{LN}(W_5 x_{m,l}^{geo}),
\end{equation}
where $W_3$, $W_4$, $W_5$ are learned projection matrices and $\text{LN}(\cdot)$ is layer normalization. We note that a 3D proposal $O_m$ corresponds to multiple 2D semantic feature vector $I_{m,l}$ obtained from different frames $l$. We randomly choose one of $I_{m,l}, l\in \{1,\cdots,L \}$ as the corresponding 2D semantics in each epoch of training in SAT. We refer to the sampled $d$-dimensional 2D semantic vector as $I_{m}$, which corresponds to the 3D proposal $O_m$.

\noindent\textbf{Text embedding.}
Given a query with $K$ words, we embed the text input with a pre-trained BERT model~\cite{devlin2018bert} into a set of $d$-dimensional word feature vectors $\{Q_1,\cdots,Q_K\}$. We fine-tune the BERT text encoder during training.

\subsection{Fusion and grounding module}
After respectively embedding each modality into multiple $d$-dimension feature vectors, we apply a stack of transformer layers~\cite{vaswani2017attention} to fuse the input modalities (query words, 3D objects proposals, and if training 2D semantics). We denote the transformer's output features at the language, 3D proposal, and 2D semantics positions as $F^{Q}$, $F^{O}$, and $F^{I}$.

An output grounding module that consists of two fully connected layers projects fused features $\{F^{O}_1,\cdots,F_M^{O}\}$ into a set of $M$ grounding scores $\{S^{O}_1,\cdots,S_M^{O}\}$, respectively. The object proposal $m$ with the highest grounding score is selected as the final grounding prediction.

\section{2D Semantics Assisted Training (SAT)}
\label{sec:sat}

SAT learns {auxiliary alignments} between the 2D object semantics and the objects in 3D scenes/language queries to assist 3D visual grounding. Figure~\ref{fig:arch} overviews SAT in training and inference with the 3D grounding transformer. 

We study two technical problems in SAT. First, in Section~\ref{sec:loss}, we propose the auxiliary training objectives that align 2D semantics with the 3D scene and language query. Second, in Section~\ref{sec:mask}, we introduce the 2D semantics encoding method that generates the fused feature $F^I$ from 2D inputs $I$. We use $F^I$ in computing the auxiliary losses.

\subsection{Training objectives}
\label{sec:loss}
In addition to the main training objective between the 3D scene and language query, SAT introduces auxiliary training objectives to align 2D semantics with the 3D scene and language query. We apply the ``visual grounding loss'' between the query and 3D/2D visual inputs. We propose an ``object correspondence loss'' between the 3D and 2D objects.

\noindent\textbf{3D visual grounding loss.} 
We first introduce the main visual grounding loss $\mathcal{L}_{VG}^{O}$ between the 3D scene and language query~\cite{luo2017comprehension,yang2019fast,achlioptas2020referit3d,chen2019scanrefer}. Visual grounding loss $\mathcal{L}_{VG}$ is a softmax loss over grounding scores $S_m^O$ for proposals $m\in \{1,\cdots,M \}$. The proposal with the highest Intersection over Union (IoU) with the ground truth region is labeled $1$ and all remaining ones have label $0$ (the highest IoU equals $1.0$ when experimented with ground truth object proposals). $\mathcal{L}_{VG}^{O}$ encourages the model to generate high scores for positive proposals. In inference, the proposal with the highest score $S^O$ is selected as the final prediction.

\noindent\textbf{2D visual grounding loss.} 
We apply the 2D grounding loss $\mathcal{L}_{VG}^{I}$ with the same form as $\mathcal{L}_{VG}^{O}$ between the 2D semantics and language query. A separate grounding head with two fully connected layers projects the fused features $\{F^{I}_1,\cdots,F_M^{I}\}$ into the 2D grounding scores $\{S^{I}_1,\cdots,S_M^{I}\}$. $\mathcal{L}_{VG}^{I}$ is the softmax loss computed over $S^I$.

\noindent\textbf{Object correspondence loss.}
The proposed object correspondence loss learns the correspondence between the objects in 3D scenes and the ones in 2D images. We design the object correspondence loss as a triplet loss~\cite{karpathy2015deep,faghri2017vse++,wang2019learning,li2019visual}: 
\begin{align*}
        \mathcal{L}_{cor} = \sum_{m=1}^M &\left\{ \left[ \alpha- s(F_m^{O},F_m^{I}) + s(F_m^{O},F_{i}^{I}) \right ]_+ \right . \\
+ & \left . \left[ \alpha- s(F_m^{O},F_m^{I}) + s(F_{j}^{O},F_m^{I}) \right ]_+ \right\},
\end{align*}
where $\text{s}(\cdot)$ is the similarity function. We use the inner product over the L2 normalized feature $F^{O}$ and $F^{I}$ as $\text{s}(\cdot)$ in our experiments. $\alpha$ is the margin with a default value of $0.1$. $i,j$ are the index for the hard negatives where $i=\text{argmax}_{i\neq m} s(F_m^{O},F_{i}^{I})$ and $j=\text{argmax}_{j\neq m} s(F_{j}^{O},F_m^{I})$. We compute the object correspondence among the 3D and 2D object proposals $m$ within each sample (3D scene). We do not construct negatives across different 3D scenes.

We optimize the model with the following loss function:
\begin{equation}
\label{eq:loss}
    \mathcal{L} = \mathcal{L}_{VG}^{O} +\mathcal{L}_{VG}^{I} + \mathcal{L}_{cor} * w_{cor} + (\mathcal{L}_{cls}^O+\mathcal{L}_{cls}^Q) * w_{cls},
\end{equation}
where $w_{cor}$ is the weight for the object correspondence loss with a default value of $10$. In addition to 3D/2D grounding loss $\mathcal{L}_{VG}$ and object correspondence loss $\mathcal{L}_{cor}$, we add query and object classification losses $\mathcal{L}_{cls}^Q$ and $\mathcal{L}_{cls}^O$ as in Referit3D~\cite{achlioptas2020referit3d}. The query feature $Q_0$ and proposal feature $O$ are projected with fully connected layers to predict the object classes for the language query and 3D proposals. We follow the classification loss weight $w_{cls}$ of $0.5$~\cite{achlioptas2020referit3d}. Ablation studies on the losses are in the supplementary material.
\subsection{2D semantics encoding}
\label{sec:mask}
SAT uses the fused 2D semantic feature $F^I$ to compute 2D visual grounding loss $\mathcal{L}_{VG}^{I}$ and object correspondence loss $\mathcal{L}_{cor}$. In this subsection, we introduce how to encode $F^I$ from 2D semantics $I$.
We show that a simple yet effective approach to encode $F^{I}$ is by introducing proper attention masks in the multi-modal transformer.
Specifically, we adopt the same stack of transformer layers to jointly encode the three input modalities $Q$, $O$, and $I$. We design the attention mask in Figure~\ref{fig:arch} (a) such that $F^{Q}$ and $F^{O}$ do not directly attend to 2D inputs $I$ (the top five rows of the mask).
In this way, the proposed mask prevents the model from directly using 2D inputs $I$ for grounding prediction and thus avoids the performance drop in inference when $I$ is not available. Meanwhile, the proposed attention mask allows the model to reference both 2D semantics $I$ and other input features $Q$ and $O$ when generating $F^{I}$ (the bottom three rows of the mask).

We find that both properties of the proposed mask, \ie, masking 2D inputs $I$ from $F^{Q}$ and $F^{O}$, and referencing $Q$ and $O$ when generating $F^{I}$, are critical to SAT's success. We discuss alternative methods as follow.
\textbf{1)} Methods that do not mask $I$ from $F^{Q}$ and $F^{O}$ will leak 2D inputs to $F^{O}$ and partially rely on $I$ to generate the grounding prediction in training. Therefore, the grounding accuracy drops catastrophically in inference when no 2D inputs $I$ are available. \textbf{2)} Encoding $F^{Q}/F^{O}$ and $F^{I}$ independently with $Q/O$ and $I$ avoid the 2D input leakage. However, without referencing the scene context $Q$ and $O$, the 2D feature $F^{I}$ fails to generate relevant object representations that effectively help 3D visual grounding. We show related ablations in Section~\ref{sec:ablation}.
\section{Experiments}

\begin{table*}[t]
\begin{minipage}{.5\textwidth}
  \centering
\centering
\caption{The 3D grounding accuracy on Nr3D~\cite{achlioptas2020referit3d} with different training data (Nr3D training set only or with extra data from Sr3D/Sr3D+).
}
\vspace{-0.0in}
\begin{tabular}{l c c c}
    \hline
    Method & Nr3D & w/~Sr3D & w/~Sr3D+ \\
    \hline
    V + L~\cite{achlioptas2020referit3d} & 26.6$\pm$0.5\% & - & - \\
    Ref3DNet~\cite{achlioptas2020referit3d} & 35.6$\pm$0.7\% & 37.2$\pm$0.3\% & 37.6$\pm$0.4\% \\
    TGNN~\cite{huang2021text} & 37.3$\pm$0.3\% & - & - \\
    {\footnotesize IntanceRefer~}\cite{yuan2021instancerefer} & 38.8$\pm$0.4\% & - & - \\
    non-SAT & 37.7$\pm$0.3\% & 43.9$\pm$0.3\% & 45.9$\pm$0.2\% \\
    SAT~(Ours) & \textbf{49.2$\pm$0.3\%} & \textbf{53.9$\pm$0.2}\% & \textbf{56.5$\pm$0.1\%} \\
    \hline
\end{tabular}
\vspace{-0.15in}
\label{table:nr3d}
\end{minipage}%
\qquad
\begin{minipage}{0.5\textwidth}
  \centering
\centering
\caption{The 3D grounding accuracy on Sr3D~\cite{achlioptas2020referit3d} with different training data (Sr3D training set only or with extra data from Nr3D).}
\vspace{-0.0in}
\begin{tabular}{l c c}
    \hline
    Method & Sr3D & w/~Nr3D \\
    \hline
    V + L~\cite{achlioptas2020referit3d} & 33.0$\pm$0.4\% & - \\
    Ref3DNet~\cite{achlioptas2020referit3d} & 40.8$\pm$0.2\% & 41.5$\pm$0.2\% \\
    TGNN~\cite{huang2021text} & 45.0$\pm$0.2\% & - \\
    IntanceRefer~\cite{yuan2021instancerefer} & 48.0$\pm$0.3\% & - \\
    non-SAT & 47.4$\pm$0.2\% & 50.1$\pm$0.1\% \\
    SAT~(Ours) & \textbf{57.9$\pm$0.1\%} & \textbf{60.7$\pm$0.2\%} \\
    \hline
\end{tabular}
\vspace{-0.15in}
\label{table:sr3d}
\end{minipage}
\end{table*}
\begin{table}
\centering
\caption{The accuracy on ScanRef~\cite{chen2019scanrefer} with different training data (ScanRef training set only or with extra data from Nr3D/Sr3D+).}
\vspace{-0.0in}
\begin{tabular}{l c c c}
    \hline
    Method & ScanRef & w/~Nr3D & w/~Sr3D+ \\
    \hline
    Ref3DNet~\cite{achlioptas2020referit3d} & 46.9$\pm$0.2\% & 47.5$\pm$0.4\% & 47.0$\pm$0.3\% \\
    non-SAT & 48.2$\pm$0.2\% & 50.2$\pm$0.1\% & 51.7$\pm$0.1\% \\
    SAT~(Ours) & \textbf{53.8$\pm$0.1\%} & \textbf{57.0$\pm$0.3\%} & \textbf{56.5$\pm$0.2\%} \\
    \hline
\end{tabular}
\vspace{-0.1in}
\label{table:scanref_gt}
\end{table}
\subsection{Datasets}
\label{sec:dataset}

\noindent\textbf{Nr3D.}
The Natural Reference in 3D (Nr3D) dataset~\cite{achlioptas2020referit3d} augments the indoor 3D scene dataset ScanNet~\cite{dai2017scannet} with $41,503$ natural language queries annotated by Amazon Mechanical Turk (AMT) workers. There exist $707$ unique indoor scenes with targets belong to one of the $76$ object classes. There are multiple but no more than six distractors (objects in the same class as the target) in the scene for each target. The dataset splits follow the official ScanNet~\cite{dai2017scannet} splits.

\noindent\textbf{Sr3D/Sr3D+.}
The Spatial Reference in 3D (Sr3D) dataset~\cite{achlioptas2020referit3d} contains $83,572$ queries automatically generated based on a ``target''-``spatial relationship''-``anchor object'' template. The Sr3D+ dataset further enlarges Sr3D with the samples that do not have multiple distractors in the scene and ends up with $114,532$ queries.

\noindent\textbf{ScanRef.}
The ScanRef dataset~\cite{chen2019scanrefer} augments the $800$ 3D indoor scenes in the ScanNet~\cite{dai2017scannet} dataset with $51,583$ language queries. ScanRef follows the official ScanNet~\cite{dai2017scannet} splits and contains $36,665$, $9,508$, and $5,410$ samples in train/val/test sets, respectively.

\subsection{Experiment settings}
\label{sec:setting}
\noindent\textbf{Evaluation metric.}
We follow the experiment settings in Referit3D~\cite{achlioptas2020referit3d} and ScanRef~\cite{chen2019scanrefer} for experiments with ground truth and detector-generated proposals, respectively.
Specifically, Referit3D~\cite{achlioptas2020referit3d} assumes the access to ground truth objects as the 3D proposals and converts the grounding task into a classification problem. The models are evaluated by the accuracy, \ie, whether the model correctly selects the referred object among $M$ proposals. We choose this ``using ground truth proposal'' setting as the default setting and present the results on all experimented datasets (Nr3d, Sr3d, and ScanRef).

Alternatively, ScanRef~\cite{chen2019scanrefer} adopts a 3D object detector~\cite{qi2019deep} to generate object proposals. On the ScanRef dataset, we also evaluate models using Acc$@k$IoU, \ie, the fraction of language queries whose predicted box overlaps the ground truth with IoU$>k$IoU. We experiment with the IoU threshold $k$IoU of $0.25$ and $0.5$. For clarity, we present the experiments with ground truth proposals in the main paper and postpone the experiments of ``SAT with detector-generated proposals'' to the supplementary material.

\noindent\textbf{Implementation details.}
We set the dimension $d$ in all transformer layers as $768$. We experiment with a text transformer with $3$ layers and a fusion transformer with $4$ layers~\cite{hu2020iterative,yang2021tap}\eat{ referencing previous vision-language models~\cite{hu2020iterative,yang2021tap}}. The text transformer is initialized from the first three layers of BERT$_\text{BASE}$~\cite{devlin2018bert}, and the fusion transformer is trained from scratch. We sample $1024$ points for each 3D proposal from its point cloud segment and encode the proposal with PointNet++~\cite{qi2017pointnet++}. We follow the max sentence length and proposal numbers in Referit3D~\cite{achlioptas2020referit3d} and ScanRef~\cite{chen2019scanrefer} when experimented on Nr3D/Sr3D and ScanRef, respectively. The model is trained with the Adam~\cite{kingma2014adam} optimizer with a batch size of $16$. We set an initial learning rate of $10^{-4}$ and reduce the learning rate by a multiplicative factor of $0.65$ every $10$ epochs for a total of $100$ epochs.

\noindent\textbf{Compared methods.}
We compare SAT with the state-of-the-art methods~\cite{achlioptas2020referit3d,chen2019scanrefer,huang2021text,yuan2021instancerefer} and the non-SAT baseline. ``\textbf{Non-SAT}'' adopts the same ``3D grounding transformer'' architecture used in ``SAT.'' The only difference is that ``non-SAT'' does not include 2D semantics in training and thus does not use the auxiliary losses $\mathcal{L}_{VG}^{I}$ and $\mathcal{L}_{cor}$.
With the same network architecture and experiment settings, ``non-SAT'' is a directly comparable baseline to ``SAT.'' The performance difference shows how much SAT could help the 3D visual grounding task.

\subsection{3D visual grounding results}
\label{sec:results}
\noindent\textbf{Nr3D.}
Table~\ref{table:nr3d} reports the grounding accuracy on the Nr3D~\cite{achlioptas2020referit3d} dataset. Both ``non-SAT'' and ``SAT'' use the 3D grounding transformer introduced in Section~\ref{sec:trans}. For SAT's reported accuracy, we encode 2D semantics $I$ from the visual feature $x^{ROI}$, object semantic feature $x^{cls}$, and geometric feature $x^{geo}$ following Eq.~\ref{eq:i}. We postpone the ablation studies on the types of 2D semantics to Section~\ref{sec:ablation}. Different columns show the results with different training data, \ie, using Nr3D's training set only, or jointly training with Sr3D/Sr3D+'s training set. We take ``SAT-Nr3D'' as the default setting and refer to it as ``SAT.'' We refer to the experiments with extra data as ``SAT w/ Sr3D/Sr3D+.''

The top five rows of Table~\ref{table:nr3d} show that our baseline ``non-SAT'' already achieves comparable performance to the state of the art (non-SAT: $37.7\%$, InstanceRefer~\cite{yuan2021instancerefer}: $38.8\%$).
By effectively utilizing 2D semantics in training, our proposed SAT improves the non-SAT baseline accuracy from $37.7\%$ to $49.2\%$, with the identical model architecture and inference inputs. SAT also outperforms the state-of-the-art accuracy~\cite{yuan2021instancerefer} of $38.8\%$ a large margin of $+10.4\%$.
Jointly using the Sr3D/Sr3D+ training data further improves the grounding accuracy. As shown in the last row, ``SAT w/ Sr3D+'' improves ``SAT-Nr3D'' from $49.2\%$ to $56.5\%$. 

Analyses reveal that SAT learns a better 3D object representation with the assist of 2D semantics, which leads to the $11.5\%$ improvement over the non-SAT baseline. The improvement brought by Sr3D/Sr3D+ mainly comes from better modeling the spatial relationships in queries. We present these analyses in Section~\ref{sec:analyses}.

\noindent\textbf{Sr3D.}
Table~\ref{table:sr3d} shows the grounding accuracy on Sr3D~\cite{achlioptas2020referit3d}. We draw similar conclusions from Table~\ref{table:sr3d} as in Table~\ref{table:nr3d} that 1) SAT significantly improves the grounding accuracy from $47.4\%$ to $57.9\%$, 2) SAT outperforms the previous state of the art~\cite{achlioptas2020referit3d,huang2021text,yuan2021instancerefer} by large margins, and 3) extra training data (Nr3D) further boosts the accuracy from $57.9\%$ to $60.7\%$.

\noindent\textbf{ScanRef.}
Table~\ref{table:scanref_gt} reports the grounding accuracy on the ScanRef dataset~\cite{chen2019scanrefer} with ground truth object proposals. We observe a significant improvement of ``SAT'' over the non-SAT baseline (SAT: $53.8\%$, non-SAT: $48.2\%$). Extra training data from Nr3D and Sr3D+ further improves the accuracy from $53.8\%$ to $57.0\%$ and $56.5\%$, respectively.

In addition to the ground-truth object proposals~\cite{achlioptas2020referit3d}, we experiment with the proposal setting in ScanRef~\cite{chen2019scanrefer} that generates proposals with a 3D detector~\cite{qi2019deep}. To apply SAT, we first compute the ground truth 2D semantics offline. In training, we match each predicted 3D proposal with a 2D semantics object that has the largest IoU with the 3D proposal. We then evaluate the models with the Acc$@0.25$ and Acc$@0.50$ metrics. SAT achieves the Acc$@0.25$ and Acc$@0.50$ of $44.54\%$ and $30.14\%$, outperforming the non-SAT baseline of $38.92\%$ and $26.40\%$ by a large margin. We introduce the details of ``SAT with detector-generated proposals'' in the supplementary material.

\subsection{Ablation studies}
\label{sec:ablation}
\noindent\textbf{Multi-modal transformer masks.}
SAT's attention mask in the multi-modal transformer has two properties, \ie, 1) masking 2D semantics $I$ from $F^{Q}$ and $F^{O}$, and 2) referencing context $Q$ and $O$ when generating $F^{I}$. We verify the importance of both properties with the ablation studies in Table~\ref{table:mask}. In training, we replace our proposed transformer's attention mask in Figure~\ref{fig:arch} (a) with the alternative masks A/B in Table~\ref{table:mask}. In inference, the model removes the extra 2D semantics input and follows the standard inference setting as in Figure~\ref{fig:arch} (b).

Mask A does not mask 2D semantics $I$ from $F^{Q}$ and $F^{O}$. We observe that the model directly relies on the extra 2D inputs $I$ for grounding prediction. Consequently, the grounding accuracy drops catastrophically to $33.9\%$ when no 2D inputs are available in inference.
Mask B encodes $F^{I}$ with 2D semantics $I$ alone. Without referencing the scene context in $Q$ and $O$, the 2D feature $F^{I}$ fails to provide a relevant object representation and is thus less effective in helping 3D visual grounding.
Although outperforming the non-SAT baseline accuracy of $37.7\%$, ``SAT-mask B'' performs much worse than the SAT with our proposed attention mask (SAT-mask B: $43.9\%$, SAT: $49.2\%$).

\begin{table}[t]
\caption{Ablation studies on 2D semantics embedding with different transformer attention masks. The gray mask color indicates prevent from attending. SAT's attention mask is in Figure~\ref{fig:arch}~(a).}
\begin{minipage}{3.4cm}
  \centering
  \includegraphics[width=3.7cm]{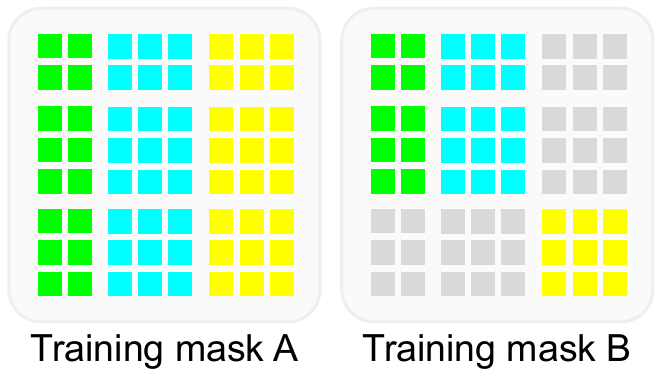}
\end{minipage}%
\qquad
\begin{minipage}{4.3cm}
    \centering
    \begin{tabular}{l c}
        \hline
        Method & Accuracy \\
        \hline
        non-SAT & 37.7$\pm$0.3\% \\
        SAT-mask A & 33.9$\pm$0.2\% \\
        SAT-mask B & 43.9$\pm$0.2\% \\
        SAT~(Ours) & \textbf{49.2$\pm$0.3\%} \\
        \hline
    \end{tabular}
    \vspace{-0.0in}
\end{minipage}%
\label{table:mask}
\end{table}

\begin{table}[t]
\centering
\caption{Ablation studies on different types of 2D semantics inputs as in Eq~\ref{eq:i}. We highlight the default ``SAT'' setting by underline.}
\vspace{-0.0in}
\begin{tabular}{l | c c c | c}
    \hline
     & +$x^{geo}$ & +$x^{cls}$ & +$x^{ROI}$ & Acc. \\
    \hline
    (a) & - & - & - & 37.7$\pm$0.3\% \\
    (b) & \cmark & - & - & 39.4$\pm$0.3\% \\
    (c) & \cmark & \cmark & - & 48.1$\pm$0.2\% \\
    (d) & \cmark & - & \cmark & 46.5$\pm$0.1\% \\
    (e) & - & \cmark & \cmark & 43.2$\pm$0.2\%\\
    (f) & \cmark & \cmark & \cmark & \underline{49.2$\pm$0.3\%} \\
    \hline
\end{tabular}
\vspace{-0.1in}
\label{table:context}
\end{table}

\noindent\textbf{Types of 2D context inputs.}
Table~\ref{table:context} shows the ablation studies on the types of 2D semantics. The combination of $x^{ROI}$, $x^{cls}$, and $x^{geo}$ are projected into a $d$-dimension 2D semantics feature $I$ following Eq.~\ref{eq:i}.

Compared with the non-SAT baseline accuracy of $37.7\%$ in row~(a), SAT with any 2D semantics significantly boosts the grounding accuracy (rows~(b-f)). Jointly using visual feature $x^{ROI}$, semantic feature $x^{cls}$, and geometric feature $x^{geo}$ achieves the best accuracy of $49.2\%$, as in row (f).
\section{How does SAT help?}
\label{sec:analyses}
\vspace{-3pt}

In this section, we analyze how does SAT help 3D visual grounding. We draw three major conclusions: \textbf{1)} SAT learns a better {3D object representation} $F^{O}$ with the assist of 2D semantics $I$, leading to consistent performance improvements over samples with different target classes, number of distractors, query lengths/types, \etc (Section~\ref{sec:probe}). \textbf{2)} Training with the extra data in Sr3D/Sr3D+ mainly benefits the queries with spatial relationship referring (Section~\ref{sec:spatial}). \textbf{3)} The performance gap between SAT and the methods that require extra 2D inputs in inference is small, \eat{directly use 2D semantics as extra inputs is small, which indicates}indicating the effectiveness of SAT in utilizing 2D semantics (Section~\ref{sec:imageinput}). Finally, we present qualitative examples in Section~\ref{sec:quali}.

\subsection{Linear probing}
\label{sec:probe}
\vspace{-3pt}
How could SAT achieve the large improvement over the non-SAT baseline and the state of the art? 
We conjecture that SAT learns a better 3D representation $F^{O}$ for noisy object point clouds with the assist of 2D semantics. Consequently, we observe consistent 3D grounding accuracy improvements on samples with different target object classes, number of distractors, query lengths/types, \etc, as shown in the performance breakdown in the supplementary material.

We use linear probing~\cite{zhang2017split,goyal2019scaling,chen2020generative} to evaluate the quality of the learned 3D object representations $F^{O}$ in different models. Specifically, we keep the pre-trained grounding network fixed and train a linear classifier that maps each proposal feature $F_m^{O}$ into one of Nr3D's $607$ object classes. \eat{We report the top-1 accuracy.}
Because no classification annotation is seen during the grounding network training, we evaluate learned representations $F^{O}$ by the object classification accuracy. Similar to the use of linear probing in representation learning~\cite{zhang2017split,goyal2019scaling,chen2020generative}, we consider a higher linear probing accuracy the indicator of a better 3D object representation $F^{O}$.

Table~\ref{table:probe} shows the linear probing accuracy on Nr3D. SAT improves the linear probing accuracy from $35.7\%$ to $60.1\%$, compared with the non-SAT baseline. The significant improvement supports the conjecture that SAT learns a better 3D object representation with 2D semantics in training\eat{to help 3D visual grounding}. We observe similar improvements in the full fine-tuning setting, where all layers are updated for object classification.
It is worth noting that SAT's effectiveness in generating better 3D object representations may hold the promise of benefiting not only 3D vision-language tasks such as grounding~\cite{achlioptas2020referit3d,chen2019scanrefer} and captioning~\cite{chen2020scan2cap}, but also 3D semantic understanding tasks such as 3D scene graph prediction~\cite{armeni20193d,wald2020learning}.

\subsection{Spatial relationship referring}
\label{sec:spatial}
\vspace{-3pt}
Our second observation is that the extra data in Sr3D/Sr3D+ helps the queries with spatial relationship referring. 
On Nr3D's subset with spatial queries ($76.7\%$ of the samples), the extra Sr3D+ training data leads to an $8.4\%$ improvement on ``SAT-Nr3D'' from $48.4\%$ to $56.8\%$. In contrast, the improvement is only $3.9\%$ on the remaining samples\eat{ without spatial relationship referring} (from $50.9\%$ to $54.8\%$). Furthermore, we observe larger improvements on subsets that contain the frequently appeared spatial words in Sr3D/Sr3D+, \eg, ``closest'' of $+11.5\%$ and ``farthest'' of $+13.5\%$.

\begin{table}[t]
\vspace{-0.23in}
\caption{Linear probing accuracy on Nr3D.}
\centering
    \begin{tabular}{l c c}
        \hline
        Method & Linear probing & Full fine-tuning \\
        \hline
        non-SAT & 35.7\% & 63.4\%\\
        SAT & 60.1\% & 65.4\% \\
        SAT w/ Sr3D+ & 61.7\% & 67.6\% \\
        \hline
    \end{tabular}
    \vspace{-0.2in}
\label{table:probe}
\end{table}

\begin{table*}[t]\small
\centering
\vspace{-10pt}
\caption{The benefit of using 2D semantics in 3D visual grounding. The upper/middle/bottom portion of the table shows the results that do not use 2D semantics/use only in training/use in both training and inference (as extra inputs). The results with extra training data (Sr3D/Sr3D+) are shown in \textcolor{gray}{gray}. Our SAT (\#(e)) shows comparable performance to oracles that require 2D inputs in inference (\#(h,i)).}
\vspace{-0.0in}
\begin{tabular}{c | l l | c c | c c c c c}
    \hline
     & \multirow{2}{*}{ } & Extra & \multicolumn{2}{|c|}{2D semantics} & \multirow{2}{*}{Overall} & \multirow{2}{*}{Easy} & \multirow{2}{*}{Hard} & \multirow{2}{*}{View-dep.} & \multirow{2}{*}{View-indep.} \\ 
     & & data & Train & Test & & & & & \\
    \hline
    (a) & Ref3DNet~\cite{achlioptas2020referit3d} & - & \xmark & \xmark & 35.6$\pm$0.7\% & 43.6$\pm$0.8\% & 27.9$\pm$0.7\% & 32.5$\pm$0.7\% & 37.1$\pm$0.8\% \\
    (b) & TGNN~\cite{huang2021text} & - & \xmark & \xmark & 37.3$\pm$0.3\% & 44.2$\pm$0.4\% & 30.6$\pm$0.2\% & 35.8$\pm$0.2\% & 38.0$\pm$0.3\% \\
    (c) & IntanceRefer~\cite{yuan2021instancerefer} & - & \xmark & \xmark & 38.8$\pm$0.4\% & 46.0$\pm$0.5\% & 31.8$\pm$0.4\% & 34.5$\pm$0.6\% & 41.9$\pm$0.4\% \\
    (d) & non-SAT & - & \xmark & \xmark & 37.7$\pm$0.3\% & 44.5$\pm$0.5\% & 31.2$\pm$0.2\% & 34.1$\pm$0.3\% & 39.5$\pm$0.4\% \\
    \hline
    (e) & SAT~(Ours) & - & \cmark & \xmark & \textbf{49.2$\pm$0.3\%} & 56.3$\pm$0.5\% & 42.4$\pm$0.4\% & 46.9$\pm$0.3\% & 50.4$\pm$0.3\% \\
    (f) & \textcolor{gray}{SAT w/ Sr3D~(Ours)} & \textcolor{gray}{Sr3D} & \cmark & \xmark & \textcolor{gray}{53.9$\pm$0.2\%} & \textcolor{gray}{61.5$\pm$0.1\%} & \textcolor{gray}{46.7$\pm$0.3\%} & \textcolor{gray}{52.7$\pm$0.7\%} & \textcolor{gray}{54.5$\pm$0.3\%} \\
    (g) & \textcolor{gray}{SAT w/ Sr3D+~(Ours)} & \textcolor{gray}{Sr3D+} & \cmark & \xmark & \textcolor{gray}{56.5$\pm$0.1\%} & \textcolor{gray}{64.9$\pm$0.2\%} & \textcolor{gray}{48.4$\pm$0.1\%} & \textcolor{gray}{54.4$\pm$0.3\%} & \textcolor{gray}{57.6$\pm$0.1\%} \\
    \hline
    (h) & 2D input aligned & - & \cmark & \cmark & 50.0$\pm$0.1\% & 62.0$\pm$0.2\% & 38.5$\pm$0.3\% & 44.7$\pm$0.3\% & 52.6$\pm$0.3\% \\
    (i) & 2D input unaligned & - & \cmark & \cmark & \underline{50.3$\pm$0.4\%} & 58.5$\pm$0.7\% & 42.4$\pm$0.5\% & 48.1$\pm$0.4\% & 51.3$\pm$0.5\% \\
    (j) & \textcolor{gray}{2D input aligned} & \textcolor{gray}{Sr3D+} & \cmark & \cmark & \textcolor{gray}{59.7$\pm$0.1\%} & \textcolor{gray}{71.0$\pm$0.3\%} & \textcolor{gray}{48.8$\pm$0.5\%} & \textcolor{gray}{52.9$\pm$0.3\%} & \textcolor{gray}{63.1$\pm$0.2\%} \\
    (k) & \textcolor{gray}{2D input unaligned} & \textcolor{gray}{Sr3D+} & \cmark & \cmark & \textcolor{gray}{61.0$\pm$0.3\%} & \textcolor{gray}{69.0$\pm$0.6\%} & \textcolor{gray}{53.2$\pm$0.3\%} & \textcolor{gray}{58.4$\pm$0.3\%} & \textcolor{gray}{62.2$\pm$0.5\%} \\
    \hline
\end{tabular}
\vspace{-0.02in}
\label{table:oracle}
\end{table*}

\begin{figure*}[t]
\begin{center}
  \centerline{\includegraphics[width=17.5cm]{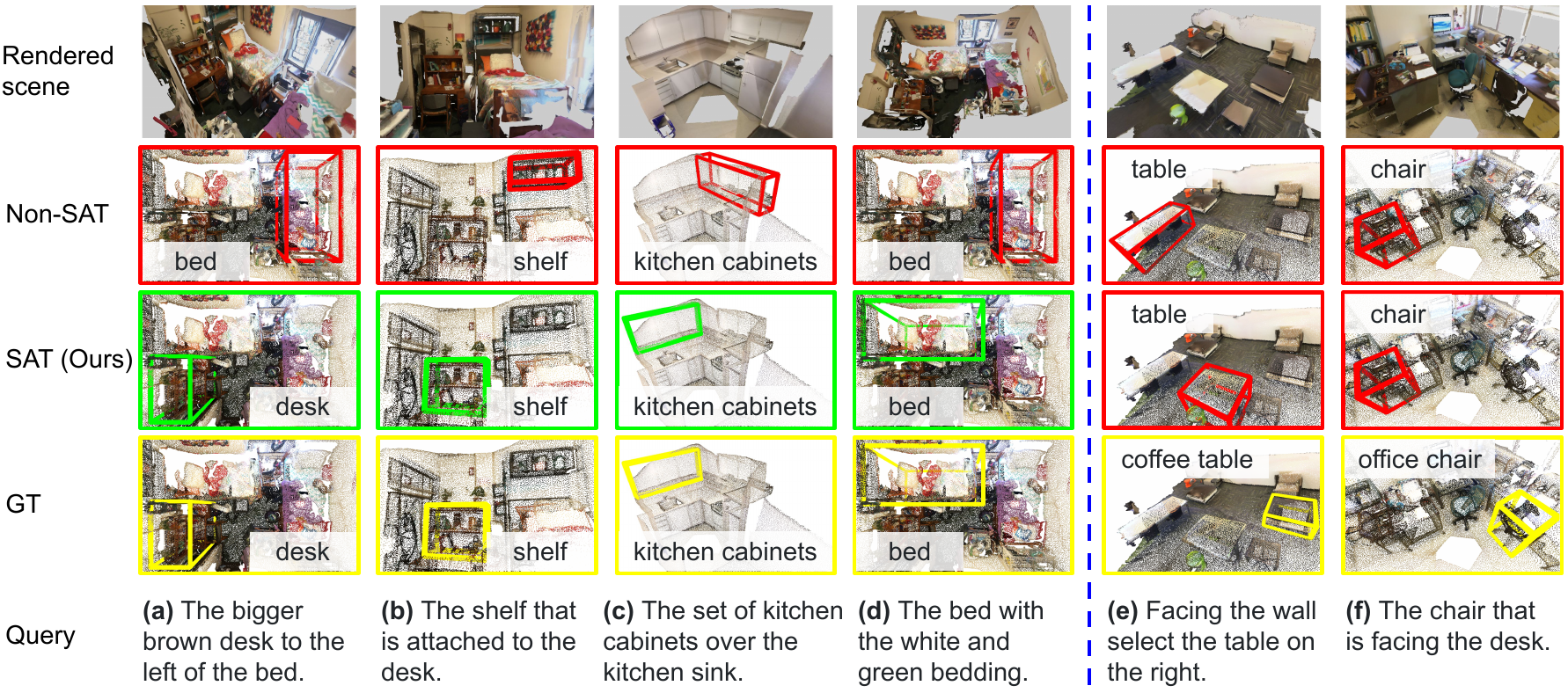}}
\end{center}
\vspace{-0.25in}
    \caption{The failure cases of non-SAT that can be corrected by SAT (the left four examples), and SAT's representative failure cases (the right two examples). The green/red/yellow colors indicate the correct/incorrect/ground truth boxes. The object class for each box is shown in text next to the 3D box. We provide rendered scenes in first row to better visualize the scene layout. Best viewed zoomed in and in color.
	}
\vspace{-0.12in}
\label{fig:visu}
\end{figure*}

\subsection{2D semantics as extra inputs}
\label{sec:imageinput}
\vspace{-3pt}

In this subsection, we compare SAT with the methods that require extra 2D inputs in both training and inference. 
We design two methods that directly use 2D semantics as extra inputs, namely the ``2D input aligned'' and ``2D input unaligned.'' Both methods use the same network architecture as SAT and take extra 2D inputs in both training and inference. For ``2D input aligned,'' we concatenate 2D semantics $I_m$ with 3D proposal feature $O_m$ and use the extended proposal feature as $O_m$ in both training and inference. The input sequence length for ``2D input aligned'' is $M+K$. We train ``2D input aligned'' with the main grounding loss $\mathcal{L}_{VG}^{O}$ and the classification loss $\mathcal{L}_{cls}$ in Eq.~\ref{eq:loss}. For ``2D input unaligned,'' we input 2D semantics $I_m$ as extra input tokens to the multi-modal transformer in both training and inference. The input sequence length for ``2D input unaligned'' is $2M+K$. We train ``2D input unaligned'' with the same loss $\mathcal{L}$ in Eq.~\ref{eq:loss} as SAT. 

Table~\ref{table:oracle} shows the experiment results on Nr3D with no 2D semantics (upper portion), with 2D inputs only in training (middle portion), and in both training and inference (bottom portion). The ``hard'' subset contains more than $2$ distractors and remaining samples belong to ``easy.''
We observe a marginal accuracy gap of $1\%$ between ``SAT'' and ``2D input aligned/unaligned'' (``overall'' in row (e): $49.2\%$, rows (h,i): $50.3\%$). The comparable performance indicates SAT's effectiveness in utilizing 2D semantics to help 3D visual grounding. Meanwhile, SAT does not require extra 2D inputs in inference as ``2D input aligned/unaligned,'' and thus is easier to use.

\subsection{Qualitative insights}
\label{sec:quali}
\vspace{-3pt}
The left four examples of Figure~\ref{fig:visu} show representative failure cases of ``non-SAT'' that can be corrected by ``SAT.'' We group common cases into three scenarios.
\textbf{1) Object:} SAT improves non-SAT by better recognizing the object classes. Non-SAT occasionally fails to ground the head noun and generates the object prediction in a different class, \eg, ``bed'' instead of the referred ``desk'' in Figure~\ref{fig:visu}~(a).
\textbf{2) Relationship:} We observe that SAT is better in modeling relationships in language queries, despite no specific modules are proposed in SAT for relationship understanding. For example, in Figures~\ref{fig:visu}~(b,c), SAT correctly understands the relationship ``attach to'' and ``over.'' We conjecture that SAT learns a better object representation for both foreground and background objects, which benefits the relationship modeling.
\textbf{3) Color and shape:} SAT also performs better in understanding color and shape-related language queries, \eg, ``white and green'' in Figure~\ref{fig:visu}~(d).

The right two examples of Figure~\ref{fig:visu} show SAT's representative failure cases. Figure~\ref{fig:visu}~(e) shows a failure case that requires understanding ``facing the wall.'' Although SAT improves both view-dependent and independent samples (\cf Table~\ref{table:oracle} ``View-dep.'' column), view understanding remains an unsolved problem. Figure~\ref{fig:visu}~(f) shows a failure case caused by ambiguous queries. The model predicts the ``\textit{chair} facing the desk'' instead of the referred ``\textit{office chair} facing the desk'' in the ground truth. We observe that the model and human annotators occasionally confuse objects in similar categories, such as chair/office-chair (Figure~\ref{fig:visu}~(f)), table/coffee-table (Figure~\ref{fig:visu}~(e)), \etc.
\section{Conclusion}
\vspace{-3pt}
We have presented 2D semantics assisted training (SAT) for 3D visual grounding. SAT uses 2D semantics in training to assist 3D visual grounding and eases joint representation learning between the 3D scene and language query. With identical network and inference inputs, SAT beats the non-SAT baseline by $11.5\%$ in absolute accuracy. SAT leads to the new state of the art on multiple datasets and outperforms previous works by large margins. Analyses show that SAT effectively uses 2D semantics to learn a better 3D point cloud object representation that helps 3D visual grounding.

{\small
\vspace{-3pt}
\subsection*{Acknowledgment}
\vspace{-4pt}
This work is supported in part by NSF awards IIS-1704337 and IIS-1813709, as well as our corporate sponsors.

\bibliographystyle{ieee_fullname}
\bibliography{egbib}
}
\clearpage
\appendix
\twocolumn[{
\begin{center}
\Large 
\textbf{SAT: 2D Semantics Assisted Training for 3D Visual Grounding}\\(Supplementary Material)
\par
\end{center}
\vspace{2em}
}]

In the first part of the supplementary material, we present additional ablation studies and detailed result analyses. In the second part, we extend our SAT approach to detector-generated 3D proposals. We also discuss how 3D proposal quality influences the 3D grounding accuracy.
\section{Experiments}

\subsection{Ablation studies}
\noindent\textbf{Training objectives.}
We conduct ablation studies on the training objectives introduced in the main paper's Eq.~2. Table~\ref{table:loss} shows the experiments on the Nr3D dataset with ground truth proposals. Same as SAT, all compared methods take extra 2D inputs in the training stage and do not require extra inputs in inference.

Row (b) shows the baseline grounding accuracy with the main 3D grounding loss $\mathcal{L}_{VG}^{O}$ and classification loss $\mathcal{L}_{cls}$ only. Despite input to the model during the training, 2D semantics $I$ does not affect the main model ($Q$ and $O$) in this baseline, as $I$ is not attended to and no $I$-related auxiliary losses are included. Therefore, row (b) is equivalent to the main paper's non-SAT baseline and shows a comparable accuracy of $38.0\%$. SAT's auxiliary objectives of 2D grounding loss $\mathcal{L}_{VG}^{I}$ and object correspondence loss $\mathcal{L}_{cor}$ improve the accuracy to $38.5\%$ and $44.9\%$, respectively, as shown in rows (c,d). Our proposed SAT jointly applies the two auxiliary objectives and achieves the best accuracy of $49.2\%$. Furthermore, we find classification loss $\mathcal{L}_{cls}$ helpful to both the baseline and the final SAT model, as shown in rows (a,b) and rows (e,f), respectively.

\subsection{Performance breakdown}
\label{sec:breakdown}
In this subsection, we show the performance breakdown on the Nr3D~\cite{achlioptas2020referit3d} dataset to better understand SAT's improvement. We report SAT's performance on subsets with different target object classes, numbers of distractors, query lengths/types, spatial relationships, \etc. We observe that SAT effectively utilizes the 2D semantics to learn better 3D object representations, and obtains consistent improvements on these subsets.
\begin{table}[t]
\centering
\caption{Ablation studies on the training objectives in the main paper's Eq.~2. Experiments are conducted on the Nr3D dataset with ground truth proposals. We highlight ``SAT'' by underline.}
\vspace{-0.0in}
\begin{tabular}{l | c c c | c}
    \hline
     & $\mathcal{L}_{cls}^O+\mathcal{L}_{cls}^Q$ & $\mathcal{L}_{VG}^{I}$ & $\mathcal{L}_{cor}$ & Acc. \\
    \hline
    (a) & - & - & - & 33.8$\pm$0.1\% \\
    (b) & \cmark & - & - & 38.0$\pm$0.3\% \\
    \rowcolor{LightGray}
    (c) & \cmark & \cmark & - & 38.5$\pm$0.3\% \\
    \rowcolor{LightGray}
    (d) & \cmark & - & \cmark & 44.9$\pm$0.2\% \\
    (e) & - & \cmark & \cmark & 46.0$\pm$0.2\% \\
    \rowcolor{LightGray}
    (f) & \cmark & \cmark & \cmark & \underline{49.2$\pm$0.3\%} \\
    \hline
\end{tabular}
\vspace{-0.0in}
\label{table:loss}
\end{table}

\begin{table}[t]\small
\centering
\caption{Grounding accuracy on Nr3D's subsets with different numbers of distractors.}
\vspace{-0.0in}
\begin{tabular}{l | c c c c c c}
    \hline
     & Overall & 2 & 3 & 4 & 5 & 6 \\
    \hline
    Percent(\%) & 100.0 & 49.0 & 21.2 & 15.9 & 8.4 & 5.5 \\
    \hline
    non-SAT & 37.6 & 44.4 & 36.8 & 25.5 & 28.8 & 28.0 \\
    SAT & 49.0 & 56.3 & 48.0 & 38.6 & 40.1 & 30.6 \\
    \hline
\end{tabular}
\vspace{-0.0in}
\label{table:numobj}
\end{table}

\begin{table}[t]\small
\centering
\caption{Grounding accuracy on Nr3D's subsets with different query lengths.}
\vspace{-0.0in}
\begin{tabular}{l | c c c c c c}
    \hline
     & Overall & 2-6 & 7-8 & 9-10 & 11-13 & 14+ \\
    \hline
    Percent(\%) & 100.0 & 21.0 & 20.5 & 19.8 & 17.5 & 21.2 \\
    \hline
    non-SAT & 37.6 & 44.2 & 38.5 & 36.8 & 35.7 & 32.4 \\
    SAT & 49.0 & 54.8 & 52.1 & 49.4 & 46.3 & 41.8 \\
    \hline
\end{tabular}
\vspace{-0.0in}
\label{table:qlen}
\end{table}

\begin{table*}[t]\small
\centering
\caption{Grounding accuracy on Nr3D's subsets with different target object classes.}
\vspace{-0.0in}
\begin{tabular}{l | c c c c c c c c c c c c}
    \hline
     & Overall & chair & table & window & door & trash can & pillow & monitor & box & shelf & picture & cabinet \\
    \hline
    Percent(\%) & 100.0 & 10.9 & 7.2 & 6.1 & 5.9 & 5.8 & 4.1 & 4.1 & 3.4 & 3.2 & 3.2 & 3.1 \\
    Avg. \#points (K) & 3.1 & 2.2 & 4.7 & 4.2 & 3.7 & 1.3 & 0.7 & 1.2 & 0.9 & 7.1 & 1.8 & 4.5 \\
    Avg. \#distractors & 3.0 & 3.6 & 3.5 & 2.6 & 2.6 & 2.9 & 3.6 & 3.9 & 3.1 & 2.6 & 3.0 & 2.7  \\
    \hline
    non-SAT & 37.6 & 30.7 & 43.2 & 39.2 & 34.7 & 40.0 & 41.7 & 38.8 & 33.5 & 38.8 & 44.7 & 38.3 \\
    SAT & 49.0 & 45.9 & 52.0 & 54.9 & 44.6 & 53.5 & 51.1 & 54.7 & 40.6 & 42.1 & 54.0 & 53.0 \\
    \hline
\end{tabular}
\vspace{-0.05in}
\label{table:class}
\end{table*}

\begin{table*}[t]\small
\centering
\caption{Grounding accuracy on Nr3D's subsets with different spatial referring keywords in language queries.}
\vspace{-0.0in}
\begin{tabular}{l | c c c | c c c c c c c}
    \hline
     & Overall & with spatial\eat{\footnote{We manually obtain filter out the samples with spatial referrings. The percentage is slightly lower than the $90.5$\cite{achlioptas2020referit3d}.}} & w/o spatial & closest & next to & on the left & on the right & corner & fa(u)rthest \\
    \hline
    Percent(\%) & 100.0 & 76.7 & 23.3 & 14.2 & 8.4 & 5.4 & 5.4 & 5.1 & 5.0 \\
    \hline
    SAT & 49.0 & 48.4 & 50.9 & 49.5 & 47.4 & 56.3 & 48.1 & 43.8 & 37.8 \\
    SAT w/ Sr3D+ & 56.4 & 56.8 & 54.8 & 61.0 & 56.1 & 62.6 & 58.0 & 53.8 & 51.3\\
    \hline
\end{tabular}
\vspace{-0.1in}
\label{table:spatial}
\end{table*}

\noindent\textbf{Numbers of distractors.}
Table~\ref{table:numobj} shows the performance on subsets with different numbers of distractors. We compare non-SAT with our SAT in the bottom part of the table. Intuitively, we observe a performance decrease when there exist more distractors in the scene, \eg, SAT's accuracy drops from $56.3\%$ to $30.6\%$ when the distractor number increases from $2$ to $6$. On the other hand, the relative improvement of SAT over the non-SAT baseline is consistent on subsets with different numbers of distractors.

\noindent\textbf{Numbers of query words.}
We also examine the influence of query length on the grounding performance to better understand the model's performance in modeling language queries. Table~\ref{table:qlen} split Nr3D into five sub-sets that are roughly the same size based on query lengths. The results show that longer queries more challenging in general. Therefore, the grounding accuracy decreases on longer queries, \eg, from $54.8\%$ to $41.8\%$, when the query length increase from less than $6$ words to more than $14$ words. Overall, SAT consistently improves the non-SAT accuracy on different subsets.

\noindent\textbf{Target's object class.}
Table~\ref{table:class} shows the performance on subsets with different target object classes. The upper part of the table shows the percentage of samples in each subset and the subsets' average number of points/distractors. The bottom part compared non-SAT with SAT on different subsets. Overall, we observe consistent improvements on subsets with different target object classes.

\noindent\textbf{Query spatial relationships.}
As overviewed in the main paper's Section~6.3, training with Sr3D/Sr3D+ mainly benefits the queries with spatial relationship referring. We manually collect the spatial relationship keywords in Nr3D and show the grounding accuracy on each generated subsets. On the left part of Table~\ref{table:spatial}, we show the overall performance on the subset with and without spatial relationship referring. We find $76.7\%$ of the queries contain at least one spatial relationship keyword, while the remaining samples do not use spatial referring. On the subset with spatial keywords, the extra Sr3D+ training data leads to an $8.4\%$ improvement on ``SAT-Nr3D'' from $48.4\%$ to $56.8\%$. In contrast, the improvement is only $3.9\%$ on the remaining samples.
The right part of Table~\ref{table:spatial} compares the performance on subsets with specific spatial keywords. We observe larger improvements on the frequently appeared spatial keywords in Sr3D/Sr3D+. For example, the accuracy improves from $49.5\%$ to $61.0\%$ on the keyword ``closest,'' and from $37.8\%$ to $51.3\%$ on the keyword ``farthest/furthest.''

\section{SAT with detector-generated proposals}
\label{sec:detector}

\begin{table*}[t]\small
\centering
\caption{3D visual grounding accuracy on ScanRef~\cite{chen2019scanrefer} with detector-generated proposals. The upper part shows results that do not require extra input in inference, and the bottom part shows methods that use extra inputs. We highlight the best performance that does not use 2D inputs by \textbf{bold}.\eat{ Oracle scores for analyses are shown in \textcolor{gray}{gray}.}
The ``unique'' subset contains samples with no distracting objects, and the remaining samples are in the ``multiple'' subset. ``(Filt.)'' in the ``proposals'' column indicates that 3D proposals are first filtered by the object class such that the model only needs to select from the proposals in the same class as the target. ``(Filt.)'' simplifies the grounding problem by using the external object label information. }
\vspace{0.01in}
\begin{tabular}{c | l | c l | c c c c c c}
    \hline
     & \multirow{2}{*}{ } & Extra &  \multirow{2}{*}{Proposals} & \multicolumn{2}{c}{Unique} & \multicolumn{2}{c}{Multiple} & \multicolumn{2}{c}{Overall} \\
     & & 2D input & & Acc$@0.25$ & Acc$@0.5$ & Acc$@0.25$ & Acc$@0.5$ & Acc$@0.25$ & Acc$@0.5$ \\
    \hline
    (a) & ScanRef~\cite{chen2019scanrefer} & \xmark & VoteNet & 60.54\% & 39.19\% & 26.95\% & 16.69\% & 33.47\% & 21.06\% \\
    (b) & IntanceRefer~\cite{yuan2021instancerefer} & \xmark & PointGroup (Filt.) & \textbf{77.13\%} & \textbf{66.40\%} & 28.83\% & 22.92\% & 38.20\% & \textbf{31.35\%} \\
    (c) & Non-SAT & \xmark & VoteNet & 68.48\% & 47.38\% & 31.81\% & 21.34\% & 38.92\% & 26.40\% \\
    (d) & SAT~(Ours) & \xmark & VoteNet & 73.21\% & 50.83\% & \textbf{37.64\%} & \textbf{25.16\%} & \textbf{44.54\%} & 30.14\% \\
    \hline
    (e) & One-stage~\cite{yang2019fast} & \cmark & None & 29.32\% & 22.82\% & 18.72\% & 6.49\% & 20.38\% & 9.04\% \\
    (f) & ScanRef~\cite{chen2019scanrefer} & \cmark & VoteNet & 63.04\% & 39.95\% & 28.91\% & 18.17\% & 35.53\% & 22.39\% \\
    (g) & TGNN~\cite{huang2021text} & \cmark & 3D-UNet & 68.61\% & 56.80\% & 29.84\% & 23.18\% & 37.37\% & 29.70\% \\
    (h) & IntanceRefer~\cite{yuan2021instancerefer} & \cmark & PointGroup (Filt.) & 75.72\% & 64.66\% & 29.41\% & 22.99\% & 38.40\% & 31.08\% \\
    \hline
\end{tabular}
\vspace{-0.15in}
\label{table:scanref}
\end{table*}

In the main paper, we focus on the ground truth proposal setting where we assume the access to $M$ ground-truth object point cloud segments as 3D proposals~\cite{achlioptas2020referit3d}. SAT is compatible with the setting that uses detector-generated proposals~\cite{chen2019scanrefer}. In this section, we present one implementation of extending SAT with detector-generated proposals. We benchmark our approach on the ScanRef dataset~\cite{chen2019scanrefer}.

\subsection{Method}
We obtain $M$ 3D proposals and their feature $O$ with a 3D object detector~\cite{qi2019deep}. It is computationally expensive to project the 3D proposals in each iteration to get the corresponded 2D semantics. Instead, we use the same method introduced in the main paper's Section~3.2 to cache the ground truth 2D image semantics $I$. The object correspondence between detector-generated 3D proposals $O_m$ and ground-truth 2D semantics $I_n$ does not naturally exist as in the ground truth 3D proposal experiments. To get the 3D-2D object  correspondence in the training stage, we compute the 3D IoU between the generated proposals $m$ and the ground truth boxes $n$ (corresponded to 2D semantics $I_n$) and pair 3D proposals with 2D semantics online by selecting the pair with the maximum IoU. We do not apply the object correspondence loss on the pairs with an IoU less than $0.5$. With the IoU computation conducted online in each epoch, our implementation supports the end-to-end optimization of the entire framework.

\subsection{Experiment results}
Table~\ref{table:scanref} shows experiment results on the ScanRef dataset~\cite{chen2019scanrefer}. The upper part of the table contains methods that do not require extra 2D inputs in inference, and the bottom part includes methods that use 2D semantics in both training and inference. The ``unique'' subset contains samples that do not have distracting objects with the same object class as the target. The remaining samples belong to the ``multiple'' subset.
We note that one previous study~\cite{yuan2021instancerefer} simplifies the grounding problem by filtering out proposals that are not in the same object class as the target. We refer to such filtered 3D proposals as ``(Filt.)'' in the ``proposals'' column.
Consequently, methods with filtered proposals show better performances on the ``Unique'' subset, which contains no distracting object in the same class as the target. The drawback is that the external object label information is required to perform such filtering.

We focus on the metrics in the \textit{``multiple''} subset, which best indicates the models' performance of 3D visual grounding~\cite{achlioptas2020referit3d}. We draw two major conclusions. \textbf{1)} SAT significantly outperforms the non-SAT baseline by effectively utilizing the 2D semantics in the training stage (SAT: $37.64\%$ and $25.16\%$, non-SAT: $31.81\%$ and $21.34\%$). \textbf{2)} SAT outperforms the state of the art~\cite{yuan2021instancerefer,huang2021text} by large margins (SAT: $37.64\%$ and $25.16\%$, InstanceRefer: $28.83\%$ and $22.92\%$).
The significant improvements over the non-SAT baseline and the state of the art indicate the effectiveness of our approach in 3D visual grounding.

Furthermore, the state-of-the-art methods~\cite{huang2021text,yuan2021instancerefer} find it helpful to replace VoteNet with other proposal generation methods, such as 3D-UNet or PointGroup~\cite{jiang2020pointgroup}. We discuss the influence of the proposal quality in Section~\ref{sec:discuss}.

\subsection{Discussion}
\label{sec:discuss}
\noindent\textbf{3D proposal quality.}
When experimented with the detector-generated 3D proposals, the final grounding accuracy is influenced by two factors, \ie, the quality of generated proposals and the main grounding objective of point-cloud-language modeling. We observe that the current 3D proposal quality is still somewhat limited\eat{, and the grounding accuracy can be boosted by simply using stronger 3D detectors}. When using VoteNet~\cite{qi2019deep} for proposal generation, ScanRef reports an oracle Acc$@0.5$ of $ 54.33\%$, where the best proposal is selected as the final prediction. Because of the imperfect proposal quality, previous studies~\cite{huang2021text,yuan2021instancerefer} find it effective to boost the grounding accuracy by simply replacing proposal generation methods~\cite{huang2021text,jiang2020pointgroup}.

Despite the large influence of proposal quality on grounding accuracy, we argue that the point-cloud-language joint representation learning is the core problem of 3D visual grounding. We expect the fast-growing 3D object detection studies to bring stronger detectors in the future, which alleviates the proposal quality problem. Therefore, in this study, we focus on the unique joint representation learning problem in 3D visual grounding, and evaluate methods with the metrics that best reflect the models' grounding performance. Specifically, we focus on ``accuracy'' when experimented with ground truth proposals, and the ``multiple'' accuracy when experimented with detector-generated proposals. 
For the former, SAT surpasses the state of the art by large margins on Nr3D ($+10.4\%$ in absolute accuracy) and Sr3D~\cite{achlioptas2020referit3d}~($+9.9\%$).
Similarly on ScanRef, SAT-GT achieves an Acc$@0.5$ of $66.01\%$, surpassing the state of the art by large margins (ScanRef-GT: $40.06\%$, InstanceRef-GT: $55.37\%$).
For the latter,  SAT significantly outperforms the state of the art as shown in Table~\ref{table:scanref}'s ``multiple'' column. In summary, SAT effectively uses 2D semantics to assist 3D visual grounding and sets the new state of the art on multiple 3D visual grounding datasets.





\end{document}